# Melody Generation using an Interactive Evolutionary Algorithm

1st Majid Farzaneh
Faculty of Media Technology and Engineering
Iran Broadcasting University
Tehran, Iran
Majid.Farzaneh91@gmail.com

2nd Rahil Mahdian Toroghi
Faculty of Media Technology and Engineering
Iran Broadcasting University
Tehran, Iran
Mahdian.t.r@gmail.com

*Abstract*— Music generation with the aid of computers has been recently grabbed the attention of many scientists in the area of artificial intelligence. Deep learning techniques have evolved sequence production methods for this purpose. Yet, a challenging problem is how to evaluate generated music by a machine. In this paper, a methodology has been developed based upon an interactive evolutionary optimization method, with which the scoring of the generated melodies is primarily performed by human expertise, during the training. This music quality scoring is modeled using a Bi-LSTM recurrent neural network. Moreover, the innovative generated melody through a Genetic algorithm, will then be evaluated using this Bi-LSTM network. The results of this mechanism clearly show that the proposed method is able to create pleasurable melodies with desired styles and pieces. This method is also quite fast, compared to the state-of-the-art data-oriented evolutionary systems.

## I. Introduction

Music is a ubiquitous, undeniable, and perhaps the most influential part of media content. It can easily facilitate, transferring of the emotion and concepts in an artistic, and delicate way. This motivates the music accompaniment with all media types and human-oriented places, such as movies, theaters, games, shops, and so on in order to bring human pleasure.

For a machine to create automatic enjoyable music it should imitate the rules, know-how, and subtleties embedded in a pleasurable melody or a famous masterpiece of an artist. This could be extracted from a rich database of artistic music being played by famous musicians and then being learned to the machine.

There are several methods introduced so far in order to generate musical melodies automatically, such as Hidden Markov Models [21, 5, 19, 7, 16, 20, 2], models based on artificial neural networks [23, 18, 8, 3, 4, 10, 14, 26], models based on the evolutionary and population-based optimization algorithms [13, 25, 11, 22], and models based on local search algorithms [6, 9]. Recently, the sequential deep neural networks especially Long Short-Term Memory (LSTM) neural networks have become prevalently used and achieved successful results generating time series sequences [15, 1, 17].

Music generation can be viewed from different aspects. One can focus on melody generation, while others can work specifically on harmony and rhythm. In terms of data, music generation methods could also be divided into note-based and signal-based methods. In former, the machine should learn music from the music sheets, while in latter the musical audio signals are learned. Music generation can also be discussed in terms of the difficulty of performing [16, 24], and the narrative [12].

The major questions in this regard are; what kind of music do we prefer to be generated by a machine? Do we need new styles to be created, or we want the machine to resort and imitate the existed music? How the quality of the generated music could be evaluated by machine? Are we able to generate new music analogous to the manuscripts of a famous musician by machine, and how this similarity could be certified?

In this study, our assumption is that human will judge the quality of the generated melody, and will give them scores. Thereafter, the human scoring will be modeled using a Bi-directional Long-Short-Term Memory (Bi-LSTM) neural network. This neural network-based system will supersede the human scoring system and will perform as a standard evaluator of the melody generated by optimization-based models. The proposed music generation system of our paper is based on Genetic algorithm, which performs as a note-based method to create melodies.

The novelty of this paper is two-fold. First, the interaction between human and machine in order to generate new meaningful and pleasurable melodies, and second is introducing a new scoring model in order to evaluate the generated melodies.

The rest of the paper is arranged as follows. In section II, the proposed method has been represented in three phases. In section III, the experimental results have been provided and then analyzed. The paper is then terminated by our conclusion in chapter IV, followed by the cited references.

## II. The Proposed Melody Generation System

The proposed method consists of three major phases. First, a Genetic algorithm (GA) is used to generate a vast spectrum of melodies, from bad ones up to pleasurable ones. The bad melodies are the ones that are made randomly when GA starts. In the second phase, the outputs of GA are given to humans to be scored from zero to 100. Then, these generated and scored melodies are trained by a Bi-LSTM neural network. This network, when trained by a sufficient amount of musical data and associated scores, would perform like a performance evaluator of the GA music generator system. In the end, the GA music which can maximize the Bi-LSTM output as the objective function is played, as the generated pleasurable melody.

## A. Phase I: Generating Training Melodies

As already mentioned, this interval involves providing the necessary training data (i.e. melodies) for the next phase. Genetic algorithm has been chosen for this task. The reason is that GA can generate a population of melodies (chromosomes) randomly, with a vast spectrum of qualities. These melodies are generated in the form of ABC notations,[1] and then at each iteration, the best melodies among the population are selected based on their similarity to the human made melodies, which are taken from the manuscripts of the most famous musicians. Therefore, a database of melodies has to be used as part of the fitness function. Then, the number of 2-gram, 3-gram, and 4-gram structures in generated melodies are computed, which also exist in the database. Then, the fitness function is calculated as,

$$Fitness(C_i) = N_2 + 10N_3 + 100N_4 \qquad (1)$$

Where $C_i$ is i$^{th}$ chromosome and $N_2$, $N_3$ and $N_4$ are the numbers of 2-grams, 3-grams, and 4-grams respectively.

If we set the database for a specific genre, the generated music will be very much similar to that specific genre, as well. To make this stronger, we calculate the probability of each character (i.e., the characters in ABC notation regime which are going to be played) in the dataset as in the following equation,

$$P(character_i) = \frac{N_i}{N_{total}} \qquad (2)$$

Where $N_i$ is the number of occurrence of $character_i$ in the entire database and $N_{total}$ is the total number of characters in the database. Hereafter, whenever GA generates new melodies, it makes characters occur according to these probabilities. For example, if note G repeated 100 times in database, and there are 5000 characters overall in the database, the probability of character G will be 0.02 and if GA goes to generate 100 characters for a random melody, 2 characters among those 100 should be G. Therefore, we lead GA makes melodies more like the database.

The crossover and mutation are applied, then on selected chromosomes (melodies) at each iteration. The **crossover** is applied as follows,

*for i=1 to D do*
  *if rand>0.5 then*
    *child(i)=best1(i);*
  *else*
    *child(i)=best2(i);*
  *end if*
*end for*

Where *D* is the chromosome length, *rand* is a uniform random number between 0 to 1, *best1* and *best2* are the selected chromosomes and *child* is a new chromosome after crossover.

The **mutation** is applied, as follows,

*for i=1 to D do*
  *if rand>0.1 then*
    *child(i)=best1(i);*
  *else*
    *child(i)=random_melody(i);*
  *end if*
*end for*

During GA generation, we save all the generated melodies for the scoring purposes and making the training data for the second phase.

The melody generation system tries to produce similar to the melodies. The execution time of melody generation is high since it should compare every generated melody with a large dataset note-by-note, and three times. What we desire is to generate new melodies which are pleasurable for the audience, furthermore to be similar to the existing melodies. To achieve this, we need another phase to be implemented.

## B. Phase II: LSTM-based Evaluating Model

To simulate the human scoring mechanism, a Bi-LSTM neural network has been used. The reason is that the melodies are somewhat a reasonable and sensible sequence of the musical notes being chained to represent a meaningful and pleasurable sensing atmosphere to the audience. This sequence justifies using recurrent neural networks. LSTM neural network is a solution to that. Bidirectional LSTM (aka Bi-LSTM) is an LSTM whose parameters represent the forward and backward correlations of the adjacent notes or frames of the musical signal, both.

For a generated melody to be qualified as pleasurable, a high score should be assigned to it by human assessment. This score could be within 0 to 100 boundary values. Therefore, we have used the melodies as input to the Bi-LSTM network and the average of the scores as the output of it. This GA based generated melodies would never be obtainable again since they are directly used as the training data for the network. The Bi-LSTM, after being sufficiently trained would be a qualified evaluation system for the quality and pleasurability of the generated musical melodies.

## C. Phase III: The Proposed Music Generation System

This phase is the same as the first one, except that the objective function being used is different. The flowchart of the proposed music generation system is depicted in figure 1.

---

[1] For more information please refer to www.abcnotation.com

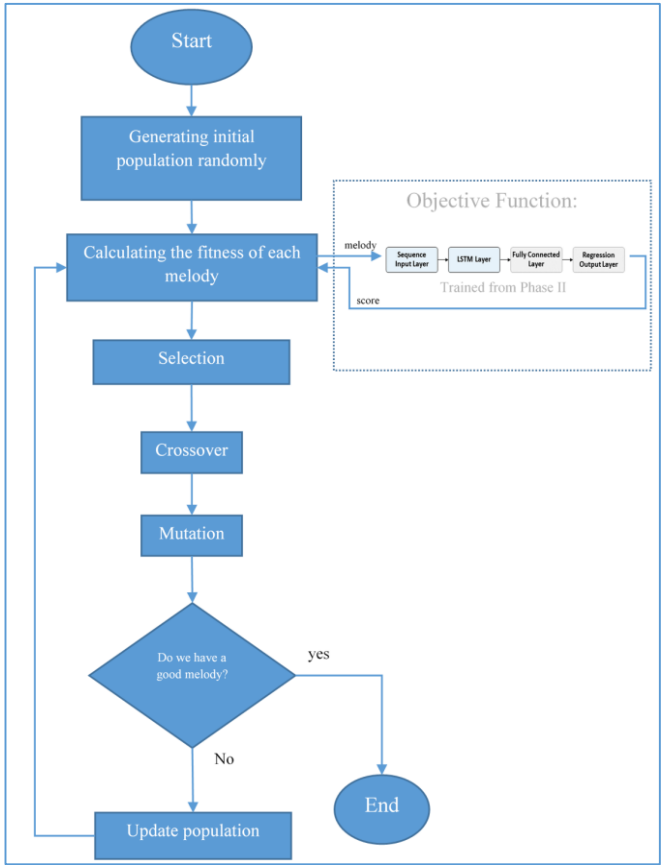

**Figure 1: The entire proposed music generation system. The GA algorithm performs as the generator system, both for training the evaluator system and for the final generation system. Using the automatic evaluator, the produced melodies are chosen among the most pleasurable ones created from the GA population.**

### III. EXPERIMENTAL RESULTS

The proposed architecture, shown in figure 1, has been implemented in MATLAB framework. For phase I, the Campin dataset [2] which contains 200 different European melodies in ABC notation has been used. Figure 2 shows the note probabilities in this dataset.

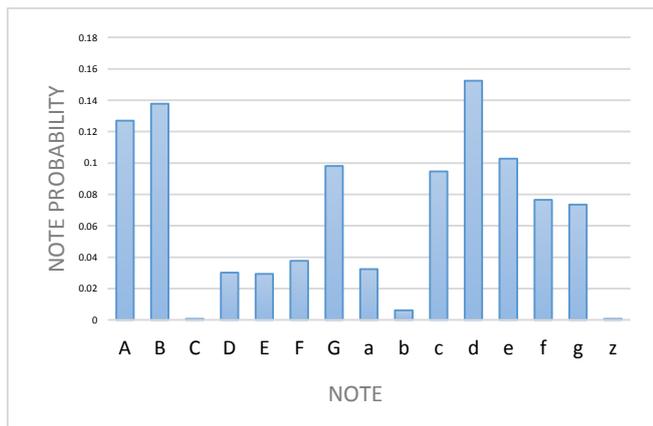

**Figure 2: Notes vs. their probability of occurrence in Campin (European melodies) dataset. Notes are represented in ABC notation standard.**

As already mentioned, the GA algorithm in phase I and III are almost the same, except for the performance function they are going to optimize which is different. The initial

---
[2] http://abcnotation.com/tunes

settings of the genetic algorithms in these two phases are depicted in table 3.

**Table 1: Initial settings**

| Parameter | Phase I | Phase III |
|---|---|---|
| Maximum iteration | 2000 | 500 |
| Population size | 20 | 20 |
| Crossover rate | 0.5 | 0.5 |
| Mutation rate | 0.1 | 0.1 |
| Fitness function | Similarity to Campin dataset | Bi-LSTM Neural Network |

The experimental settings associated with the Bi-LSTM neural network, are as follows:
- The Bi-LSTM hidden layer size is 50
- Number of epochs for the training phase is 5000
- The cost function being used in Bi-LSTM training is the mean square of error (mse)
- Melody size for both phases (I and III) contain 30 notes

During the human evaluation, we generated 6 melodies through phase I, and 6 melodies through phase III, and then we asked 20 audiences to give these melodies their scores within 0 to 100. These 12 melodies are available on YouTube to listen using the link below:

https://www.youtube.com/watch?v=Ci6DHEwAYcQ&feature=youtu.be

Figure 3 shows the mean of the scores taken from these 20 audiences and the standard deviation of the scores for each melody.

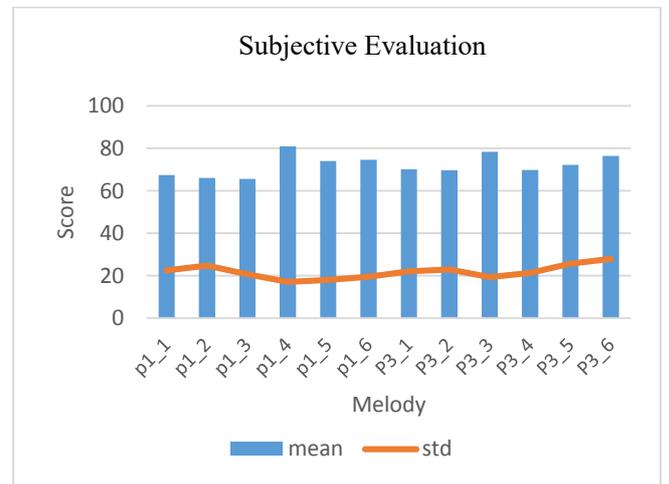

**Figure 3: Subjective evaluation of the pleasurability of the melodies, scored by 20 audiences**

Even though the genetic algorithms of phase I, and phase III are pretty similar, the execution time of these two phases differ drastically. For the training phase, the execution time is quite dependent on the number of notes it is going to generate, whereas in generation phase (Phase III) the execution time is almost independent of the number of notes it is going to generate and it remains constant. The figure clearly shows this property. This could be explained due to the heavy search the algorithm is going to do note-by-note during the training phase, as explained before.

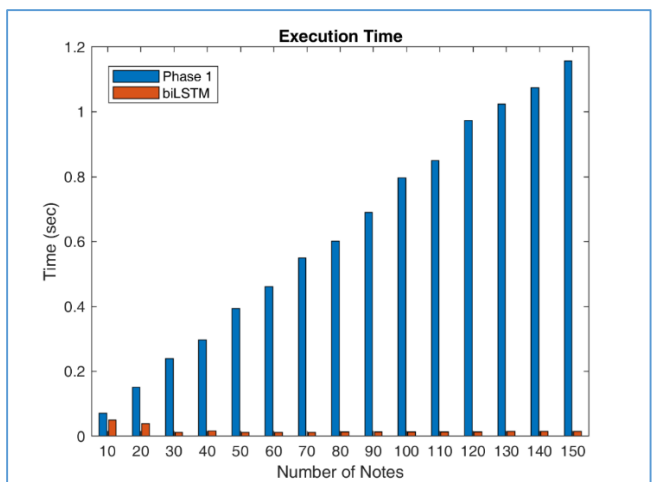

Figure 4: **The execution time at phase I, and phase III (Bi-LSTM) vs. the number of notes**

IV. CONCLUSION

In this paper, the problem of automatic music generation based on interactive (human and machine) evolutionary optimization method (here Genetic Algorithm) has been proposed. The goal of the architecture is to enable a machine to generate melodies which are pleasurable, and meaningful. Initially, the evaluation of what machine generates is performed by the listeners, however, when a Bi-LSTM neural network learns ow the human evaluates the pleasurable property of the melodies, it supersedes the human and does the task independently. Finally, the GA algorithm generates a population of melodies, among which those with highest scores (which are evaluated by the neural network) are chosen to be released. The experiments have been performed on Campin dataset, and the results have been satisfying (pleasurable and consistent) for the audiences in a subjective evaluation study. The generation process has been fairly fast, which makes this method outperform the data-based evolutionary systems.